\documentclass{article} 
\usepackage{iclr2026_re-align_workshop,times}

\usepackage{amsmath,amsfonts,bm}

\def\eqref#1{equation~\ref{#1}}

\def\1{\bm{1}}

\DeclareMathAlphabet{\mathsfit}{\encodingdefault}{\sfdefault}{m}{sl}
\SetMathAlphabet{\mathsfit}{bold}{\encodingdefault}{\sfdefault}{bx}{n}

\usepackage{hyperref}
\usepackage{url}
\usepackage{subcaption}
\usepackage{xcolor}
\usepackage{tikz}
\usepackage{tabularx}
\usepackage{booktabs}
\usepackage{wrapfig}
\newcommand{\ok}{\textcolor{green!60!black}{\large\checkmark}}
\newcommand{\bad}{\textcolor{red!70!black}{\large\texttimes}}

\usetikzlibrary{positioning,arrows.meta}

\title{The Illusion of Latent Generalization: Bi-directionality and the Reversal Curse}

\iclrfinalcopy

\author{
Julian Coda-Forno\thanks{ Work done as a student researcher in GDM.} \\
TUM, Helmholtz Munich\\
\texttt{julian.coda-forno@helmholtz-munich.de} \\
\And
Jane X. Wang \\
Google DeepMind \\
\AND
\centerline{Arslan Chaudhry} \\
\centerline{{Google DeepMind}} \\
}

\begin{document}

\maketitle

\begin{abstract}
The \textit{Reversal Curse} describes a failure of autoregressive language models to retrieve a fact in reverse order (e.g., training on ``$A > B$'' but failing on ``$B < A$''). Recent work shows that objectives with bidirectional supervision (e.g., bidirectional attention or masking-based reconstruction for decoder-only models) can mitigate the reversal curse. We extend this evaluation to include a vanilla masked language modeling (MLM) objective and compare it to decoder-only masking-based training across four reversal benchmarks and then provide a minimal mechanistic study of \emph{how} these objectives succeed. We show that reversal accuracy requires training signal that explicitly makes the source entity a prediction target, and we find little evidence that success corresponds to a single direction-agnostic representation of a fact. Instead, representation distances and linear probes are consistent with storing forward and reverse directions as distinct entries, with different indexing geometry for MLM versus decoder-only masking-based training. Our results caution that objective-level ``fixes'' can improve reversal behavior without necessarily inducing the kind of latent generalization one might expect from a unified concept.
\end{abstract}

\section{Introduction}
\label{sec:intro}

Large Language Models (LLMs) can generalize flexibly from information provided \emph{in-context} \cite{Lampinen2025Generalization}, yet often fail to use knowledge learned \emph{in the weights} with the same flexibility---a brittleness recently framed as the \textit{latent generalization} problem \cite{lampinen2025latentlearningepisodicmemory}. A clean symptom is the \textit{Reversal Curse} \cite{Berglund2024Reversal}: a model trained on a fact in one direction (e.g., ``$A > B$'') can fail to retrieve the logically equivalent reverse (``$B < A$'').

Several mitigation strategies exist. Data-level interventions (e.g., explicit reverse augmentation) can work but may distort language statistics or require manual paraphrase generation \cite{Golovneva2024Reverse,Lampinen2025Generalization}. A more fundamental approach modifies the \emph{training objective} to provide bidirectional supervision, either explicitly via bidirectional attention \cite{lv-etal-2024-analysis, nie2025largelanguagediffusionmodels} or implicitly for decoder-only models via masking-based post-training \cite{pan2025closingdataefficiencygapautoregressive}. These objectives can close the \emph{behavioral} gap, but it remains unclear \emph{how} they do so mechanistically: do they produce a direction-agnostic representation of a fact, or do they instead learn a second, separately-indexed memory entry for the reverse query?

In this paper, we ask what changes \emph{inside} a model when the reversal curse is fixed. We begin by confirming a robust behavioral pattern: objectives that provide bidirectional supervision -- explicitly via masked language modeling (\textbf{MLM}), or implicitly via masking-based post-training for decoder-only models (\textbf{NTP+Masking}) -- achieve non-zero reversal accuracy in settings where standard next-token prediction (NTP) collapses (Fig.~\ref{fig:mlm_vs_ntp}). We then probe what signal is actually doing the work. First, we isolate which prediction targets are necessary for reversal success (Table~\ref{tab:ablation1_zero}, Fig.~\ref{fig:masking_variants_joint}). Second, we analyze the learned representations to test whether success corresponds to a single, direction-agnostic ``fact'' representation or to storing an additional entry specialized for the reverse query (Figs.~\ref{fig:reps}--\ref{fig:probes}). 

Overall, our results point to the latter interpretation: both successful objectives behave as if forward and reverse directions are stored as distinct entries, but the geometry of how these entries are indexed differs between MLM and NTP+Masking.

\section{Setup}
\label{sec:setup}

\subsection{Problem Setup \& Notation}
\label{sec:notation}
We formalize a fact as a tuple $(s, r, t)$ consisting of a source entity $s$, a relation $r$, and a target entity $t$. In natural language, this is expressed as a sequence $s \xrightarrow{r} t$. Throughout, we use ``$A > B$'' to denote a forward fact with source $A$ and target $B$, and ``$B < A$'' to denote its linguistic reverse. Entities $A,B$ may be multi-token spans (e.g., names like \textit{Daphne Barrington}). We always refer to the entity `A' as source and entity `B' as target regardless of their position (as subject or target) in the forward or reverse direction. 

\paragraph{Evaluation Protocol (Reversal Accuracy).}
For each benchmark, most facts are shown in \emph{both} directions during training to teach that the relation is reversible. For a held-out subset of test facts, the model is trained \emph{only} on the forward direction (e.g., ``$A > B$'') and evaluated on the reverse prompt (e.g., ``$B < \dots$''). We report the fraction of test prompts (\textbf{accuracy}) for which the model outputs the correct source entity (in reverse direction).

\subsection{Data}
\label{sec:data_brief}
We evaluate on four reversal benchmarks spanning synthetic to semantic structure: \textbf{Simple Reversal} \cite{lampinen2025latentlearningepisodicmemory}, \textbf{Nonsense Entities} \cite{Lampinen2025Generalization}, \textbf{Fictional Celebrity} \cite{Berglund2024Reversal}, and \textbf{Semantic Structure} \cite{Lampinen2025Generalization}. Full templates, split ratios, and augmentation details are provided in App.~\ref{app:data}.

\subsection{Models and Training}
\label{sec:models_eval}

To isolate the effect of finetuning vs pretraining, we use two settings.

\paragraph{Training from scratch (Simple Reversal).}
For \textit{Simple Reversal}, we train models from scratch using the transformer architecture described in \citet{lampinen2025latentlearningepisodicmemory}. We compare a decoder-only NTP model against an encoder-style MLM variant (no causal mask).

\paragraph{Pretrained models (language-based datasets).}
For \textit{Nonsense Entities}, \textit{Fictional Celebrity}, and \textit{Semantic Structure}, we compare a pretrained BERT-Large (340M)~\citep{Devlin2019BERT} for \textbf{MLM} against Gemma-3 4B~\citep{team2025gemma} for decoder-only objectives: for decoder-only masking-based training (\textbf{NTP+Masking}), we follow \citet{pan2025closingdataefficiencygapautoregressive} and use the instruct-tuned Gemma-3 4B variant; for standard \textbf{NTP}, we use the base Gemma-3 4B.

\subsubsection{Training Objectives}
\label{sec:objectives_main}

We evaluate three training objectives:
\begin{enumerate}\setlength{\itemsep}{2pt}
    \item \textbf{NTP:} standard next-token prediction with a causal mask.
    \item \textbf{MLM:} masked language modeling with bidirectional attention \cite{Devlin2019BERT}.
    \item \textbf{Masked Fine-Tuning (NTP+Masking).}
    Following \citet{pan2025closingdataefficiencygapautoregressive}, we train a decoder-only model on examples where a \emph{masked} version of a passage is placed in the context before the \emph{unmasked} passage, and the loss is computed on the unmasked continuation. For a fact ``$A > B$'', an example is:
    \texttt{[MASK] > B [SEP] A > B}.
    Intuitively, the masked context encourages the model to infer missing tokens bidirectionally, while still training with standard next-token prediction on the unmasked segment. At test time, we provide no special tokens and evaluate on the standard reverse prompt ``$B <$''.
\end{enumerate}

\section{Results}
\label{sec:results}

\subsection{Bidirectional supervision fixes the reversal curse}
\label{sec:replication}

\begin{wrapfigure}{r}{0.52\linewidth}
  \vspace{-6pt}
  \centering
  \includegraphics[width=\linewidth]{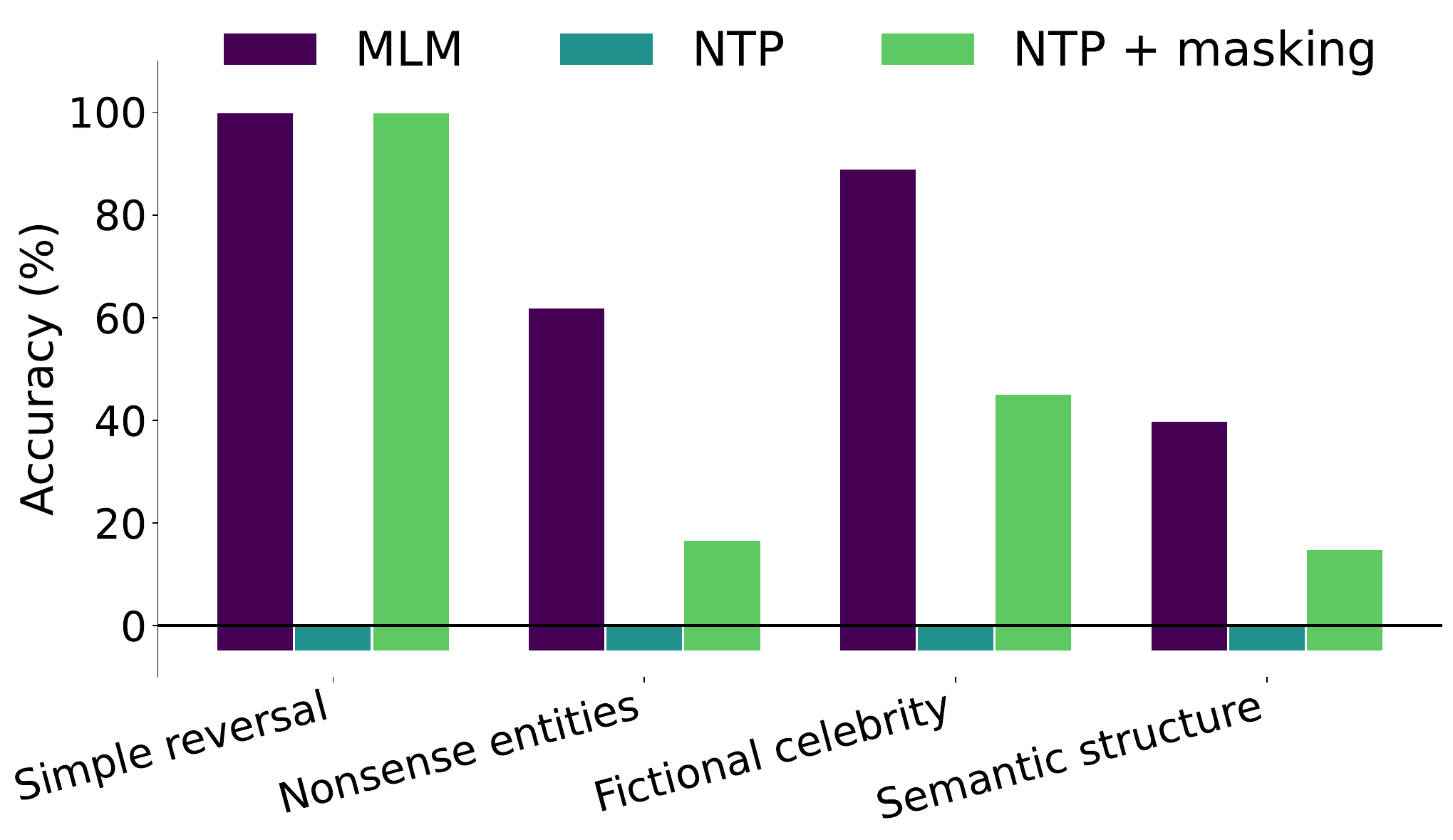}
  \vspace{-16pt}
  \caption{Reversal accuracy across datasets. Standard NTP models fail (near 0\%), while MLM and
decoder-only masking-based training achieve robust reversal accuracy, indicating that bidirectional
supervision (explicit or implicit) is sufficient to mitigate the reversal curse behaviorally.}
  \label{fig:mlm_vs_ntp}
  \vspace{-30pt}
\end{wrapfigure}

We first establish the baseline performance difference between NTP, MLM, and NTP+Masking across our benchmarks. We evaluate on held-out test sets requiring retrieval in reverse order (Train: ``$A > B$'', Test: ``$B < \dots$'').

Figure~\ref{fig:mlm_vs_ntp} replicates the core behavioral claim: providing bidirectional supervision closes the reversal gap. However, this does not yet explain \emph{what} is learned internally---nor whether MLM and NTP+Masking rely on the same mechanism.

\subsection{When does masking help? Source prediction is necessary (but not always sufficient)}
\label{sec:ablation1}

A reversal query requires predicting the \emph{source} from the \emph{target} plus relation. This motivates a direct test: does reversal require training signal on \(p(\text{source} \mid \text{relation}, \text{target})\)?

\paragraph{Ablation: never predict the source.}

\begin{wraptable}{r}{0.56\linewidth}
  \vspace{-8pt}
  \centering
  \small
  \setlength{\tabcolsep}{4pt}
  \renewcommand{\arraystretch}{1.10}
  \begin{tabular}{@{}lccc@{}}
    \toprule
    \textbf{Dataset} & \textbf{MLM} & \textbf{NTP+Mask} & \textbf{Abl.\ I} \\
    \midrule
    Simple Reversal & 100\% & 99.5\% & \textbf{0\%} \\
    Nonsense        & 60.2\% & 16.7\% & \textbf{0\%} \\
    Celebrity       & 86.4\% & 45.0\% & \textbf{0\%} \\
    \bottomrule
  \end{tabular}
  \vspace{-6pt}
\caption{Ablation I: removing source prediction (never masking $A$) collapses reversal accuracy to 0\% across datasets, despite non-zero reversal accuracy under the standard MLM / NTP+Masking setups (Fig.~\ref{fig:mlm_vs_ntp}).\\}
  \label{tab:ablation1_zero}
  \vspace{-10pt}
\end{wraptable}
We modify training so that the \textbf{source entity tokens ($A$) are never masked}---hence are never predicted. We still allow masking of relation tokens and the target $B$. We then evaluate reversal accuracy as usual. This collapses reversal accuracy to \textbf{0\%} on all datasets where the ablation is well-defined (Table~\ref{tab:ablation1_zero}), showing that reversal success depends on ever requiring the model to predict the source given the target and relation.

\paragraph{Masking sweep (Simple Reversal).}
To isolate what drives success, we focus on \textit{Simple Reversal}---where both MLM and ``NTP+Masking'' achieve $100\%$ under their standard configurations (Table~\ref{tab:ablation1_zero}). We then run a \emph{masking sweep}: we vary which components are ever masked (and thus prediction targets)—the \textit{source} $A$, relation token, and/or \textit{target} $B$—and measure reversal accuracy for each variant.

\begin{figure*}[ht!!!!!]
    \centering
    \begin{tabular}{@{}c c@{}}
        \textbf{NTP + Masking} & \textbf{MLM} \\
        \includegraphics[width=0.48\textwidth]{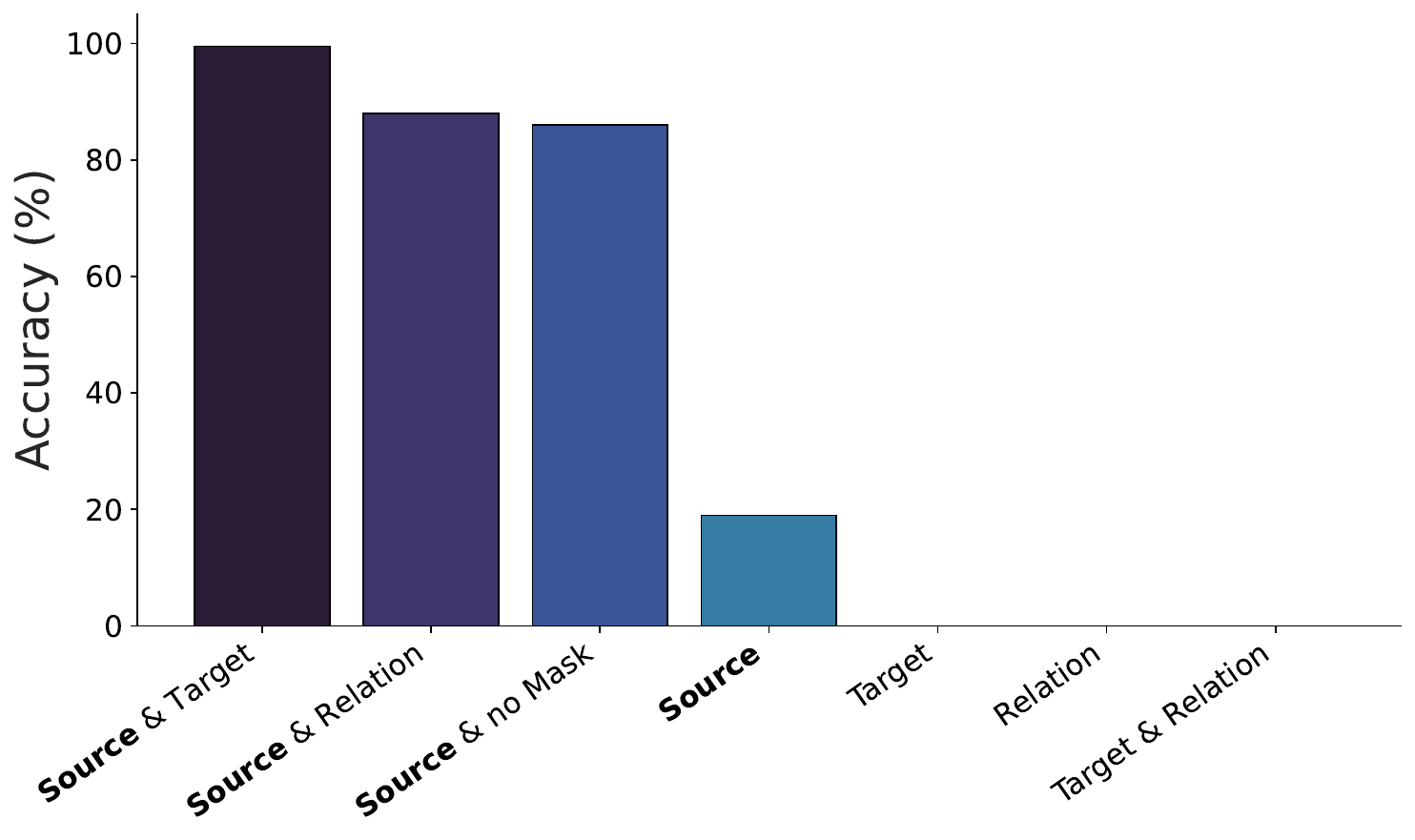} &
        \includegraphics[width=0.48\textwidth]{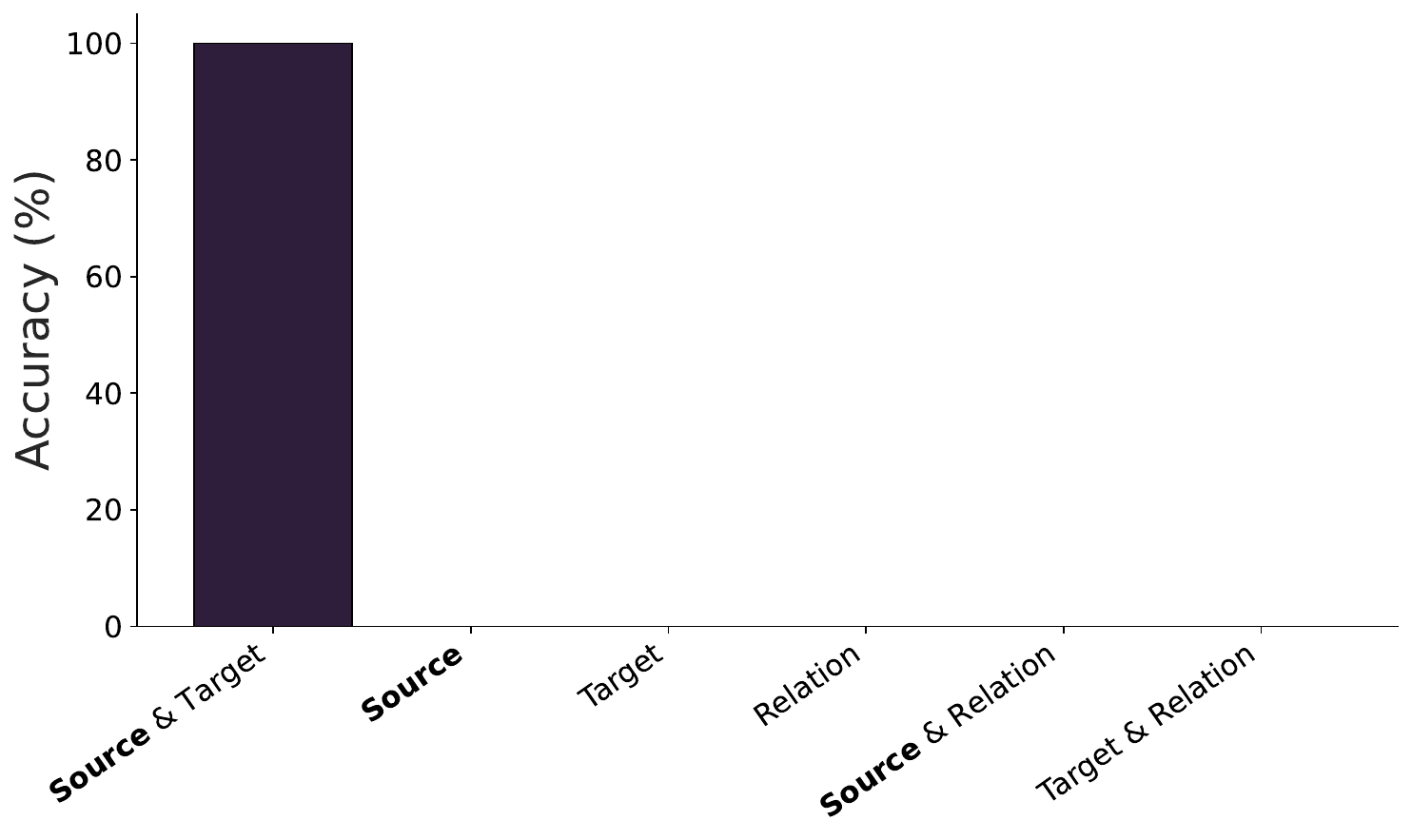} \\
        
    \end{tabular}
    \caption{\textbf{Ablation: masking sweep on Simple Reversal:} Reversal accuracy as a function of which components are masked (and therefore prediction targets) during training. \textbf{Left (NTP + Masking):} masking the \emph{source} is necessary; when the source is never masked, accuracy drops to 0\%. \textbf{Right (MLM):} masking the source alone is not sufficient; only masking \textbf{Source \& Target} reliably succeeds.}
    \label{fig:masking_variants_joint}
\end{figure*}

Figure~\ref{fig:masking_variants_joint} summarizes the sweep. For NTP+Masking (right), making the source $A$ a prediction target is close to sufficient: whenever $A$ is never masked, reversal collapses to 0\%, whereas any variant that sometimes masks $A$ yields non-zero performance. For MLM (left), the condition is stricter: masking the source alone does not reliably succeed, and the only consistently successful variant masks \textbf{Source \& Target}. Thus, even though both objectives require some training signal that targets the source, they differ in how much additional supervision is needed for that signal to generalize to the evaluation query.

\subsection{Representation analysis: reversal without a unified concept}
\label{sec:repr}

Behaviorally, MLM and NTP+Masking both mitigate the reversal curse (Fig.~\ref{fig:mlm_vs_ntp}). Mechanistically, we ask whether this reflects a direction-agnostic latent concept (forward and reverse co-localized), or two separately-stored entries that merely support the reverse query.

\begin{figure*}[ht!!!!!!]
    \centering
    \textbf{A} \hspace{8cm} \textbf{B} \\
    \includegraphics[width=0.95\textwidth]{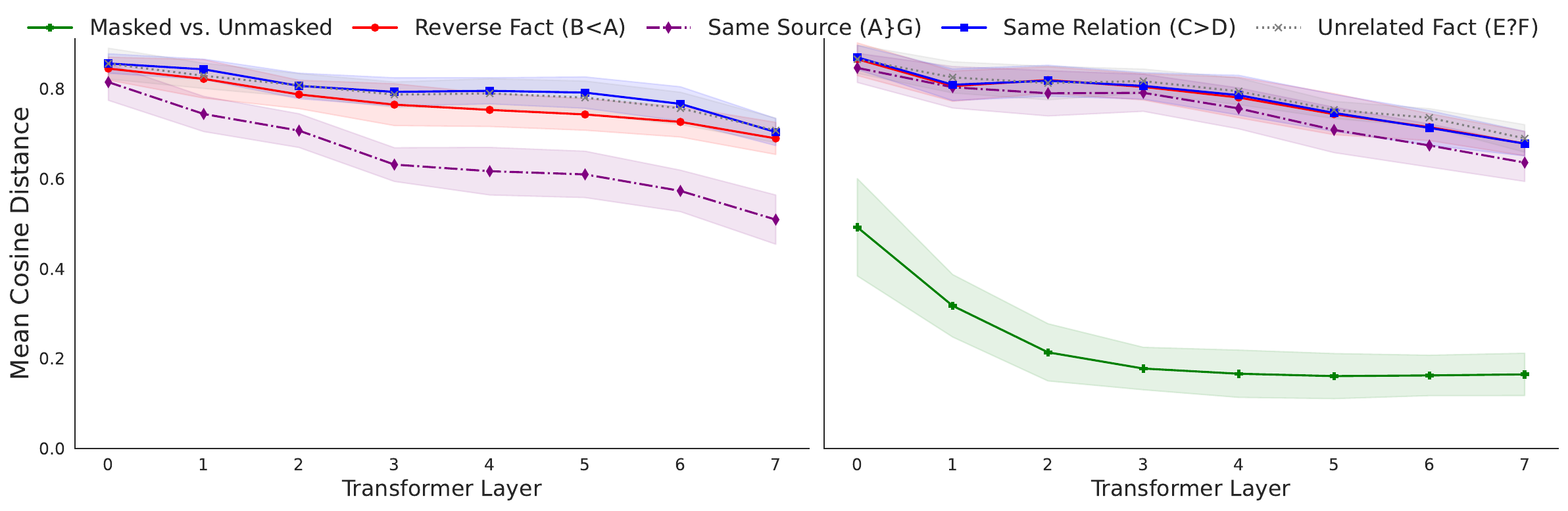}
    \caption{\textbf{Representational Distance in MLP Hidden Layers}. Mean Cosine Distance across transformer layers between a fact representation (e.g., $A > B$) and related/unrelated facts for \textbf{(A)} NTP + Masking and \textbf{(B)} MLM. The reverse fact ($B < A$) remains far from the forward fact, suggesting the two directions are stored as distinct entries rather than a unified direction-agnostic concept.} 
    \label{fig:reps}
\end{figure*}

\subsubsection{Relative neural distance.}
We analyze internal representations within the MLP blocks, which can function as key-value memories \cite{geva2021transformerfeedforwardlayerskeyvalue}. We extract MLP-post-nonlinearity states at the \emph{answer slot}: for NTP/NTP+Masking, the state used to predict the missing entity (``$A > \dots$''$\to B$, ``$B < \dots$''$\to A$); for MLM, the state at \texttt{[MASK]} (``$A >$ \texttt{[MASK]}'' and ``$B <$ \texttt{[MASK]}'' ).
 MLM provides an internal control by comparing masked vs.\ unmasked states of the same fact (green curve in Fig.~\ref{fig:reps}B). This "Masked vs. Unmasked" baseline represents the minimum achievable distance for the "same" fact, a reference point not natively available in the decoder-only setup. We measure the \textit{Mean Cosine Distance} between these states and reference points averaged across 20 facts. We use cosine distance specifically because the angular orientation of vectors has been shown to encode stable semantic structures in neural networks \cite{mikolov2013efficient, wang2024tracing}. 

\textbf{Result.} In both objectives, the reverse fact ($B < A$) remains far from the forward representation (Fig.~\ref{fig:reps}), inconsistent with a simple representational ``collapse'' into a direction-agnostic concept. Instead, both are consistent with ``two-slot'' storage. Importantly, the detailed structure differs: NTP+Masking shows stronger subject-centric clustering (same-source facts are closer), while MLM shows weaker structure (reverse resembles unrelated facts).

\subsubsection{Linear probing for inseparability.}



The significant distances in Fig.~\ref{fig:reps} suggest that the model's internal geometry does not treat a fact and its reverse as privileged counterparts. However, distance alone does not preclude a more subtle possibility: the two directions might be far apart, yet still be connected by a \emph{consistent mapping}. Concretely, the model might encode a stable transformation that takes the representation of a forward fact to the representation of its reverse---a ``semantic bridge'' that could implement reversal as a reusable operation rather than as an additional lookup.

This idea is motivated by a classic observation in high-dimensional embedding spaces: relational structure can manifest as consistent offset vectors (e.g., $King - Man \approx Queen - Woman$). If bidirectional objectives truly induce a unified, direction-agnostic understanding of facts, we would expect the forward$\rightarrow$reverse relationship to be special in this sense: the vector that connects a fact to its reverse should have a characteristic signature that differs from vectors connecting that fact to arbitrary other memories.

\begin{wrapfigure}{r}{0.50\linewidth}
  \vspace{-6pt}
  \centering
  \includegraphics[width=\linewidth]{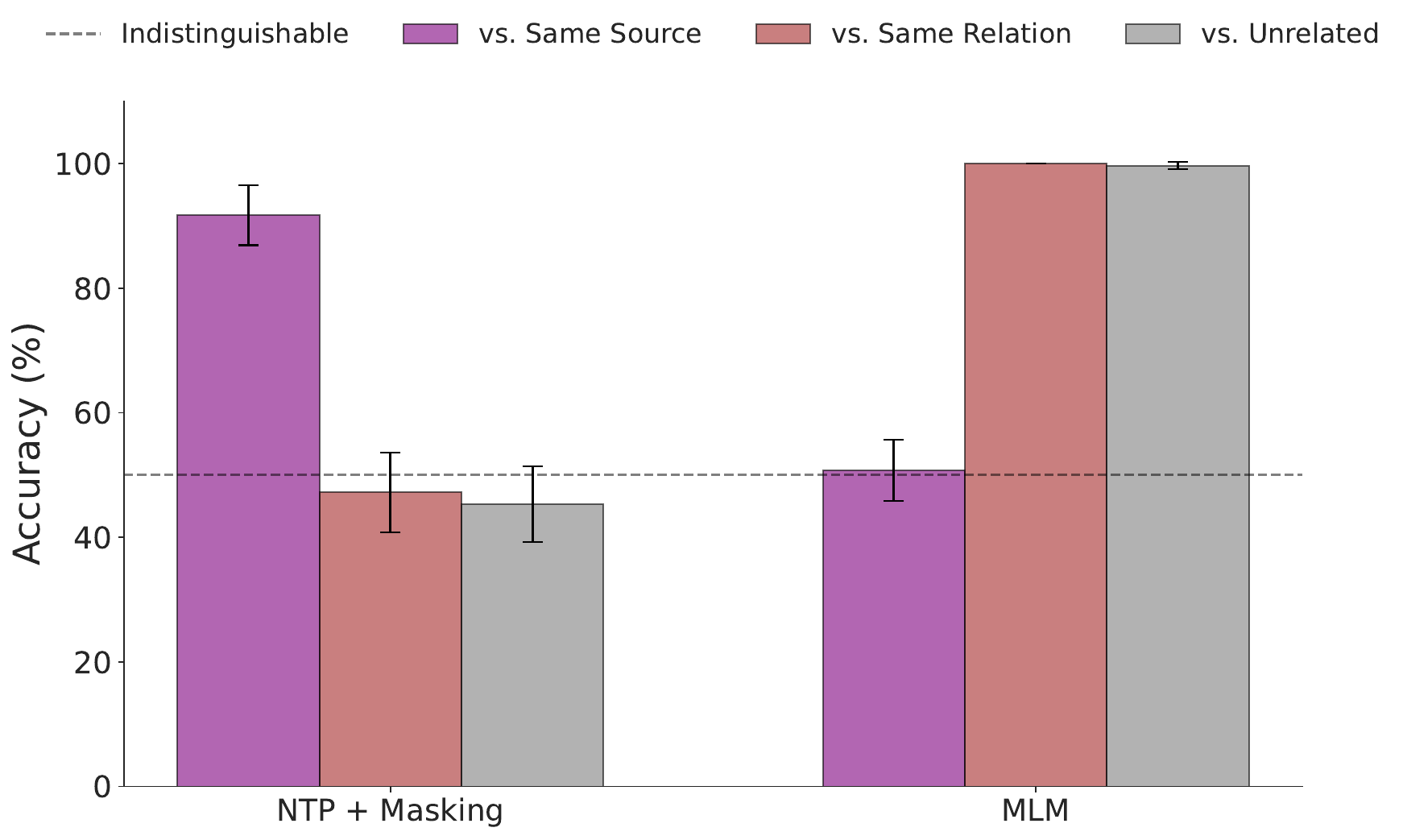}
  \vspace{-4pt}
    \caption{\textbf{Linear inseparability of reversals}. Accuracy of a logistic regression probe trained to distinguish $(Fact_1 - ReverseFact_1)$ from $(Fact_1 - Fact_2)$. In NTP+Masking, reversals are indistinguishable from unrelated facts; in MLM, they are primarily indistinguishable from same-source facts.}
  \label{fig:probes}
  \vspace{5pt}
\end{wrapfigure}

We test this directly using \emph{difference vectors}. For each held-out fact, we compute
\[
\Delta_{\text{rev}} \;=\; Fact_1 - ReverseFact_1,
\]
and compare it to
\[
\Delta_{2} \;=\; Fact_1 - Fact_2,
\]
where $Fact_2$ is chosen from one of three control sets: \textit{Same Source}, \textit{Same Relation}, or \textit{Unrelated}. We then train a linear probe (logistic regression) to classify whether a given difference vector came from a true reversal pair ($\Delta_{\text{rev}}$) or from a non-reversal comparison ($\Delta_2$). 

A key point is how to interpret failure. If the probe achieves only chance accuracy ($\approx 50\%$), then $\Delta_{\text{rev}}$ is \emph{linearly indistinguishable} from $\Delta_{2}$: the forward$\rightarrow$reverse ``direction'' is no more recognizable to a linear readout than the direction from a fact to some other stored entry. Mechanistically, this means there is no simple, reusable vector-level signature for ``reverse this fact'' encoded in the representation space. In that case, reversal behavior is more naturally explained by \textit{symmetric encoding}: the reverse direction is encoded as an additional entry, rather than derived via a privileged transformation from the forward one.

Figure~\ref{fig:probes} shows that this is exactly what happens, but in \emph{different ways} for the two objectives. For NTP+Masking, $\Delta_{\text{rev}}$ is already indistinguishable from \textit{Unrelated} (and \textit{Same Relation}) comparisons: the reverse direction looks, to a linear readout, like a jump to arbitrary knowledge. For MLM, $\Delta_{\text{rev}}$ is primarily indistinguishable from \textit{Same Source} comparisons: reversals are confused with other facts indexed under the same subject. Thus, both objectives solve reversal without an explicit ``reverse'' transform in representation space, but MLM appears to impose a stronger entity-centric organization than NTP+Masking. This conclusion is specific to \emph{linear} structure and does not rule out a potential systematic \emph{nonlinear} reversal mapping.

\section{Discussion and Conclusion}
\label{sec:discussion}

Our results separate \emph{behavior} from \emph{mechanism}. Bidirectional supervision (explicit MLM or decoder-only masking-based training) reliably fixes the behavioral reversal curse (Fig.~\ref{fig:mlm_vs_ntp}), but our analyses suggest this does not correspond to a single direction-agnostic representation of a fact (Figs.~\ref{fig:reps}--\ref{fig:probes}). We do not require exact order-invariant representations: natural-language reversals can differ pragmatically, and inverse relation tokens may be arbitrary. However, if bidirectional supervision induced a unified, direction-agnostic ``fact'' representation, we would still expect the two directions to become \emph{more} closely coupled in representation space than unrelated facts. Instead of a single direction-agnostic representation, both objectives are consistent with learning two distinct entries that support forward and reverse queries. Crucially, they organize these entries differently: MLM exhibits stronger subject-centric clustering, while NTP+Masking yields a geometry where reversals can be as indistinguishable as unrelated facts.

\textbf{Conclusion.}
Mitigating the reversal curse through bidirectional supervision is a meaningful step toward more data-efficient learning, but it may not resolve the underlying brittleness of latent generalization. A direct direction for future work is to design objectives or architectures that \emph{couple} the two directions of a fact---enforcing representational linkage (shared keys, explicit transforms, or consistency constraints) rather than learning a second independently-indexed memory entry.

\bibliographystyle{plainnat}

\bibliography{iclr2026_conference}

\clearpage

\appendix

\section{Dataset details}
\label{app:data}

\begin{figure*}[t]
\centering
\begin{tikzpicture}[
  font=\small,
  box/.style={draw, rounded corners, align=left, inner sep=7pt, text width=0.95\textwidth},
  method/.style={draw, rounded corners, align=left, inner sep=7pt, text width=0.95\textwidth},
  arrow/.style={-Latex, thick},
]

\node[box] (top) {
  \textbf{Held-out fact (train forward only):}\;
  \texttt{\textcolor{blue}{A} \;>\; \textcolor{orange}{B}}
  \hfill
  \textbf{Test query (reverse):}\;
  \texttt{\textcolor{orange}{B} \;<\; \dots} $\rightarrow$ predict \texttt{\textcolor{blue}{A}}
};

\node[method, below=6mm of top] (ntp) {
  \textbf{NTP (causal mask)}\par\vspace{2pt}
  \textbf{Train input:}\; \texttt{\textcolor{blue}{A} > \textcolor{orange}{B}}.\par
  \textbf{Loss:}\; next-token prediction on the sequence.\par\vspace{2pt}
  \textbf{Examples:}\par
  \hspace*{1.2em}\ok\; \texttt{\textcolor{blue}{A} > \textcolor{orange}{B}}\;\; (standard forward sequence)\par
  \hspace*{1.2em}\bad\; \texttt{\textcolor{orange}{B} < \dots}\;\; (we do \emph{not} train on reverse prompts)
};

\node[method, below=4mm of ntp] (mlm) {
  \textbf{MLM (bidirectional attention)}\par\vspace{2pt}
  \textbf{Train input:}\; corrupt by masking tokens.\par
  \textbf{Ablation 1 examples:}\par
  \hspace*{1.2em}\ok\; \texttt{\textcolor{blue}{A} \; \colorbox{gray!20}{[MASK]} \; \textcolor{orange}{B}}\par
  \hspace*{1.2em}\ok\; \texttt{\textcolor{blue}{A} > \colorbox{gray!20}{[MASK]}}\par
  \hspace*{1.2em}\bad\; \texttt{\colorbox{gray!20}{[MASK]} > \textcolor{orange}{B}}
  \;\; (disallowed: would mask \texttt{\textcolor{blue}{A}})\par
  \textbf{Loss:}\; predict the masked token(s).
};

\node[method, below=4mm of mlm] (ntpm) {
  \textbf{NTP + Masking (causal mask)}\par\vspace{2pt}
  \textbf{Train input:}\; masked context + unmasked continuation.\par
  \textbf{Ablation 1 examples:}\par
  \hspace*{1.2em}\ok\; \texttt{\textcolor{blue}{A} > \colorbox{gray!20}{[MASK]} \; [SEP] \; \textcolor{blue}{A} > \textcolor{orange}{B}}\par
  \hspace*{1.2em}\ok\; \texttt{\textcolor{blue}{A} \; \colorbox{gray!20}{[MASK]} \; \textcolor{orange}{B} \; [SEP] \; \textcolor{blue}{A} > \textcolor{orange}{B}}\par
  \hspace*{1.2em}\bad\; \texttt{\colorbox{gray!20}{[MASK]} > \textcolor{orange}{B} \; [SEP] \; \textcolor{blue}{A} > \textcolor{orange}{B}}
  \;\; (disallowed: masks \texttt{\textcolor{blue}{A}} in context)\par
  \textbf{Loss:}\; next-token prediction only on the unmasked continuation segment (everything after [SEP]).
};

\draw[arrow] (top.south) -- (ntp.north);

\end{tikzpicture}
\caption{Overview of training setups.}
\label{fig:setup_overview_rows_badges}
\end{figure*}
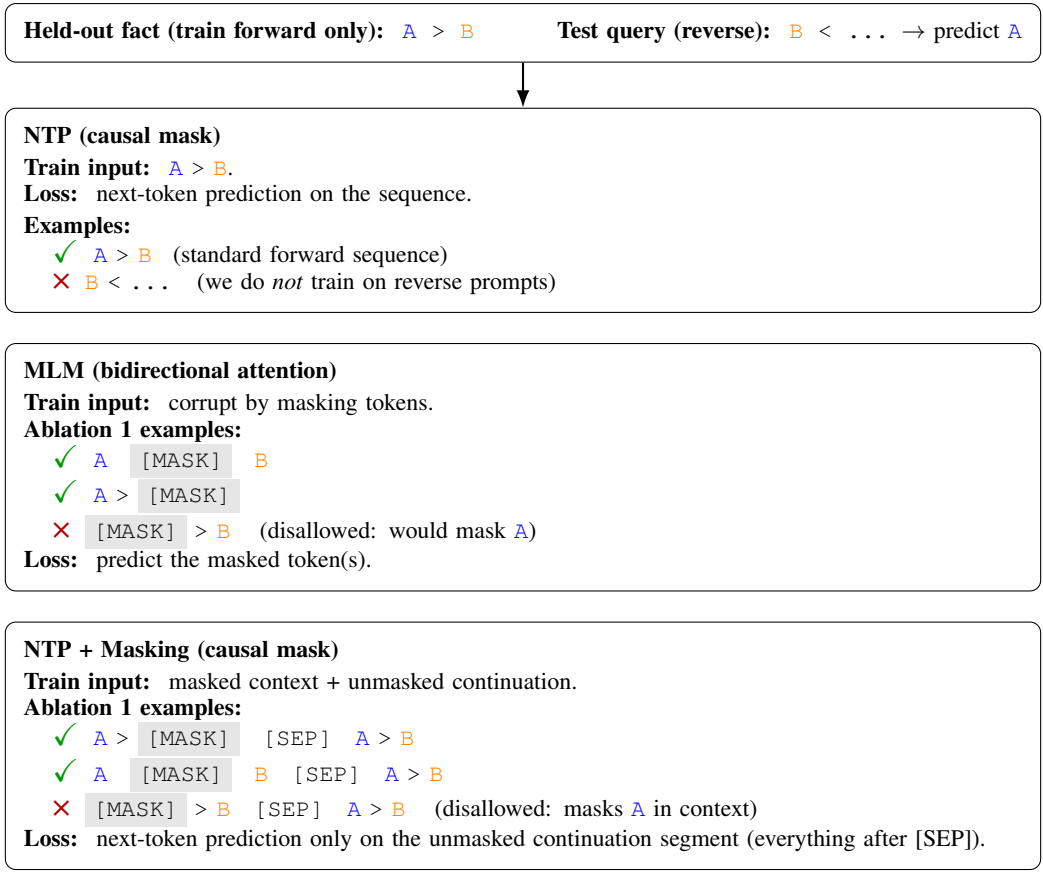
 
\subsection{Datasets}
To ensure the robustness of our findings, we utilize four distinct datasets ranging from synthetic to semantic knowledge. Importantly, all datasets are novel, i.e. not contaminated from pre-training.

\paragraph{Simple Reversal \citep{lampinen2025latentlearningepisodicmemory}:}
This dataset is explicitly designed to have sufficient diversity for minimal training from scratch. The universe consists of 1,000 entities and 20 relations (plus a reverse relation for each).
\begin{itemize}
    \item \textbf{Data Volume:} The total universe of facts is $1,000 \text{ entities} \times 20 \text{ relations} \times 2 \text{ directions} = 40,000$ sequences.
    \item \textbf{Split:} We hold out 200 specific facts for testing. For these 200 facts, the model is trained only on the forward sequences. The remaining  39,800 sequences are included in the training set to establish the structural patterns of the relations.
    \item \textbf{Augmentation:} To create diverse training documents, each relation sentence is augmented with up to 5 random prefix and suffix tokens.
\end{itemize}

\paragraph{Nonsense Entities \citep{Lampinen2025Generalization}:}
This dataset consists of synthetic nonsense words constructed to test structural learning without any potential semantic leakage from pre-training.
\begin{itemize}
    \item \textbf{Example:} ``\textit{femp} are more dangerous than \textit{glon}.''
    \item \textbf{Structure:} There are 100 such comparisons provided, with comparison words sampled from a set of 28 (e.g., ``brighter'', ``heavier''). Each comparison is repeated across 10 different training examples, paired with randomly sampled preambles.
    \item \textbf{Split:} To facilitate structural learning, 40\% of the forward sentences in the training data are explicitly augmented with their reverse counterparts. The test set consists of the remaining facts where the reverse direction was strictly held out.
\end{itemize}

\paragraph{Fictional Celebrity \citep{Berglund2024Reversal}:}
This dataset consists of fictional celebrity names and their descriptions.
\begin{itemize}
    \item \textbf{Example:} ``\textit{Daphne Barrington} is the director of \textit{'A Journey through Time'}.''
    \item \textbf{Split details:} 50 \%.
\end{itemize}

\paragraph{Semantic Structure \cite{Lampinen2025Generalization}:}
This benchmark employs a relational semantic hierarchy with nonsense terms to test deductive inference.
\begin{itemize}
    \item \textbf{Example:} ``\textit{gruds}'' (dogs) are a type of ``\textit{abmes}'' (mammals).
    \item \textbf{Structure:} The hierarchy includes 110 categories with naturalistic, asymmetric properties. To mitigate tokenization artifacts, nonsense terms are formed from plausible English phoneme combinations (4-5 letters).
\end{itemize}

\section{Objective details and hyperparameters}
\label{app:objectives}

\paragraph{Optimization.}
Unless stated otherwise, we fine-tune pretrained models with learning rate $2\times 10^{-5}$ for $10^5$ steps. For \textit{Simple Reversal}, we train from scratch using the architecture and training setup in \citet{lampinen2025latentlearningepisodicmemory}.

\paragraph{MLM masking rates.}
For BERT-style MLM, we use the standard masking procedure: each token is selected for masking with probability $0.15$ \cite{Devlin2019BERT}. (For multi-token entities and natural language prompts, this yields diverse corrupted contexts.)

\paragraph{Decoder-only masking rates (NTP + Masking).}
For NTP+Masking on language-based datasets, we follow \citet{pan2025closingdataefficiencygapautoregressive} and sample a masking ratio uniformly from $(0.05, 0.95)$ per example, then mask tokens accordingly in the context segment placed before the unmasked target segment.

\paragraph{Special case: Simple Reversal masking.}
In \textit{Simple Reversal}, each fact is exactly three tokens ($A$, relation, $B$). Rather than applying a 15\% token-level rate (which would rarely mask anything), we construct controlled variants by masking \emph{exactly one} of the three positions (source, relation, or target), or pairs thereof, matching the sweep in Fig.~\ref{fig:masking_variants_joint}.

\subsection{Illustration of training samples on \textit{Simple Reversal}}
\label{app:objective_examples}

Table~\ref{tab:objective_examples} illustrates, for a single fact ``$A > B$'', what training sequences look like under each objective (for \textit{Simple Reversal}). This is only meant as an intuition pump; for multi-token entities, MLM uses token-level masking with probability 0.15 and NTP+Masking uses a sampled masking ratio.

\begin{table}[t]
\centering
\footnotesize
\setlength{\tabcolsep}{6pt}
\renewcommand{\arraystretch}{1.15}
\begin{tabularx}{\columnwidth}{@{}lX@{}}
\toprule
\textbf{Objective} & \textbf{Example training sequences for fact $A > B$} \\
\midrule
NTP &
\texttt{A > B} \\
\midrule
MLM &
\texttt{[MASK] > B},\;
\texttt{A [MASK] B},\;
\texttt{A > [MASK]}
\\
\midrule
NTP + Masking &
\texttt{A > B},\;
\texttt{[MASK] > B [SEP] A > B},\;
\texttt{A [MASK] B [SEP] A > B},\;
\texttt{A > [MASK] [SEP] A > B}
\\
\bottomrule
\end{tabularx}
\caption{Illustrative training sequences for \textit{Simple Reversal}. MLM predicts masked tokens directly. NTP+Masking places a masked context before the unmasked continuation and trains the model to reconstruct the unmasked segment.}
\label{tab:objective_examples}
\end{table}

\section{Auxiliary ablations}
\label{app:aux}

\subsection{Ablation II: Probing for Representational Collapse}
Given that the successful models require predicting $A$ in the context of $B$ to succeed, we investigated the nature of this link. Do these models learn a structured relationship, or does the bidirectional attention (or its approximation) simply cause a ``representational collapse'' where $A$ and $B$ become strongly associated in a bag-of-words manner, ignoring syntax?

If the representations had collapsed (i.e., the model simply learns ``$A$ goes with $B$''), we would expect the model to predict $A$ regardless of the syntactic frame. We tested this by probing the successful MLM and ``NTP + Masking'' models with the novel structure: ``$B > ... ?$''. Note that this is a false statement in our setup (as $B$ is the object), but if the model had merely associated the entities, the presence of $B$ should trigger the retrieval of $A$ even in this incorrect frame. Table~\ref{tab:ablation2_examples} lists representative probe instances.

\begin{table*}[t]
    \centering
    \footnotesize
    \setlength{\tabcolsep}{6pt}
    \renewcommand{\arraystretch}{1.12}
    \begin{tabularx}{\textwidth}{@{}
        >{\raggedright\arraybackslash}p{0.14\textwidth}
        >{\raggedright\arraybackslash}X
        >{\raggedright\arraybackslash}X
        >{\centering\arraybackslash}p{0.07\textwidth}
    @{}}
        \toprule
        \textbf{Dataset} &
        \textbf{Training fact example} &
        \textbf{Probe (false frame; relation unchanged)} &
        \textbf{Acc.} \\
        \midrule
        Simple Reversal &
        $A > B$ &
        $B > \text{[MASK]}$ \; (would wrongly suggest \textcolor{red}{$A$}) &
        \textbf{0\%} \\
        Nonsense &
        ``femp are \textbf{more dangerous than} glon'' &
        ``glon are \textbf{more dangerous than} \text{[MASK]}'' \; (would wrongly suggest \textcolor{red}{femp}) &
        \textbf{0\%} \\
        \bottomrule
    \end{tabularx}
    \caption{Ablation II (syntax probe): if the model had collapsed to a bag-of-entities association, the target ($B$ / ``glon'') would trigger predicting the training-time source (in red) even in an incorrect syntactic frame. Instead, accuracy remains 0\% on these probes.}
    \label{tab:ablation2_examples}
\end{table*}

We report this probe for \textit{Simple Reversal} and \textit{Nonsense}. We omit \textit{Celebrity} because that dataset does not train/test on an explicit inverse relational operator: many prompts are effectively definitional (``\emph{X is the director of Y}'' $\rightarrow$ ``\emph{Who is X?}''), so the ``wrong-frame'' construction does not cleanly correspond to flipping the syntactic direction while keeping the same relation token. We also omit \textit{Semantic Structure} for the same reason as above: the hierarchical templates make it ambiguous what constitutes a comparable single-slot ``false frame'' probe.

The models achieved \textbf{0\% accuracy} on this probe (Table~\ref{tab:ablation2_examples}). This result indicates that the representations have \textit{not} collapsed. The models distinguish the contexts: they retrieve $A$ when $B$ appears in a reverse frame (or in the ``[MASK] $>$'' slot), but correctly refuse to associate them in the wrong syntactic direction. Given this, one could argue that MLM is in fact doing latent learning behaviorally. Whether it does so by encoding forward and reverse facts in separate memories is an implementation debate.

\subsection{Ablation III: The Prediction Bias}
Finally, we investigated the limits of what information an MLM model actually encodes. In Ablation I, we observed that predicting the source is necessary for reversal. This raises a fundamental question about latent generalization: If a model attends to a token constantly but is never forced to predict it, does it learn that token at all?

To test this, we trained models on the forward sequences ``$A > B$''. We used a modified MLM objective where we masked $A$ and the relation tokens, but \textbf{strictly never masked $B$}. We then tested the model on the \textit{forward} prediction: ``$A > \text{[MASK]}$'' (expecting $B$). Note that this is not a reversal task; it tests the model's ability to recall the exact sequence seen during training.

Despite $B$ being fully visible in the attention mechanism during every training step (while predicting $A$ or $>$), the model achieved \textbf{0\% accuracy} when asked to predict $B$ at test time.

This reveals a severe \textbf{Prediction Bias}: the model fails to learn representations for tokens that are not explicitly targets of the loss function. Even though $B$ is required to predict $A$, the model treats $B$ merely as a conditioning context and does not encode it as a generatable output. This suggests a bleak outlook for latent generalization in current architectures: information encoded solely in activations (via attention) without a corresponding gradient update on the token itself is effectively invisible to the generation mechanism.

\paragraph{Connection to post-training.}
This offers a lens on a standard post-training choice: many SFT/RLHF pipelines compute loss only on the \emph{answer} tokens, not on the \emph{prompt/question}. This is often desirable (it discourages parroting the prompt), but it may also reinforce the same bias: prompt-side spans can condition behavior without being learned as reliably generatable outputs when those spans are later required.

\end{document}